\documentclass[10pt,twocolumn,letterpaper]{article}

\usepackage{cvpr}              

\usepackage{graphicx}
\usepackage{amsmath}
\usepackage{amssymb}
\usepackage{booktabs}
\usepackage{gensymb}
\usepackage{multirow}

\usepackage[pagebackref,breaklinks,colorlinks]{hyperref}

\usepackage[capitalize]{cleveref}
\crefname{section}{Sec.}{Secs.}
\Crefname{section}{Section}{Sections}
\Crefname{table}{Table}{Tables}
\crefname{table}{Tab.}{Tabs.}

\def\cvprPaperID{6848} 
\def\confName{CVPR}
\def\confYear{2023}

\begin{document}

\title{DETR4D: Direct Multi-View 3D Object Detection with Sparse Attention}

\author{ \;
Zhipeng Luo$^{1,2}$ \;
Changqing Zhou$^{2}$\;
Gongjie Zhang$^{1}$ \; 
Shijian Lu$^1$\thanks{Corresponding author} \\
$^1$ Nanyang Technological University \;
$^2$ Sensetime Research \\
}
\maketitle

\begin{abstract}
3D object detection with surround-view images is an essential task for autonomous driving. In this work, we propose DETR4D, a Transformer-based framework that explores sparse attention and direct feature query for 3D object detection in multi-view images. We design a novel projective cross-attention mechanism for query-image interaction to address the limitations of existing methods in terms of geometric cue exploitation and information loss for cross-view objects. In addition, we introduce a heatmap generation technique that bridges 3D and 2D spaces efficiently via query initialization. Furthermore, unlike the common practice of fusing intermediate spatial features for temporal aggregation, we provide a new perspective by introducing a novel hybrid approach that performs cross-frame fusion over past object queries and image features, enabling efficient and robust modeling of temporal information. Extensive experiments on the nuScenes dataset demonstrate the effectiveness and efficiency of the proposed DETR4D. 
\end{abstract}

\section{Introduction}
\label{sec:intro}
3D object detection is a fundamental perception task, which is of great value to many important tasks, such as autonomous driving and robotics. While LiDAR-based approaches~\cite{yan2018second, shi2019pointrcnn, lang2019pointpillars, zhou2018voxelnet, yin2021centerpoint} have demonstrated outstanding performance in recent years, camera-only detection frameworks have also attracted extensive attention, not only out of cost-effective considerations but also due to the advantageous properties of images, including high resolution and rich semantic information. 
However, while LiDAR-based methods leverage native 3D geometric cues from point clouds, generating 3D bounding boxes from only images is ill-posed due to the absence of depth information, which poses unique challenges to camera-only 3D detection algorithms.

\begin{figure*}
    \centering
    \includegraphics[width=1.0\linewidth]{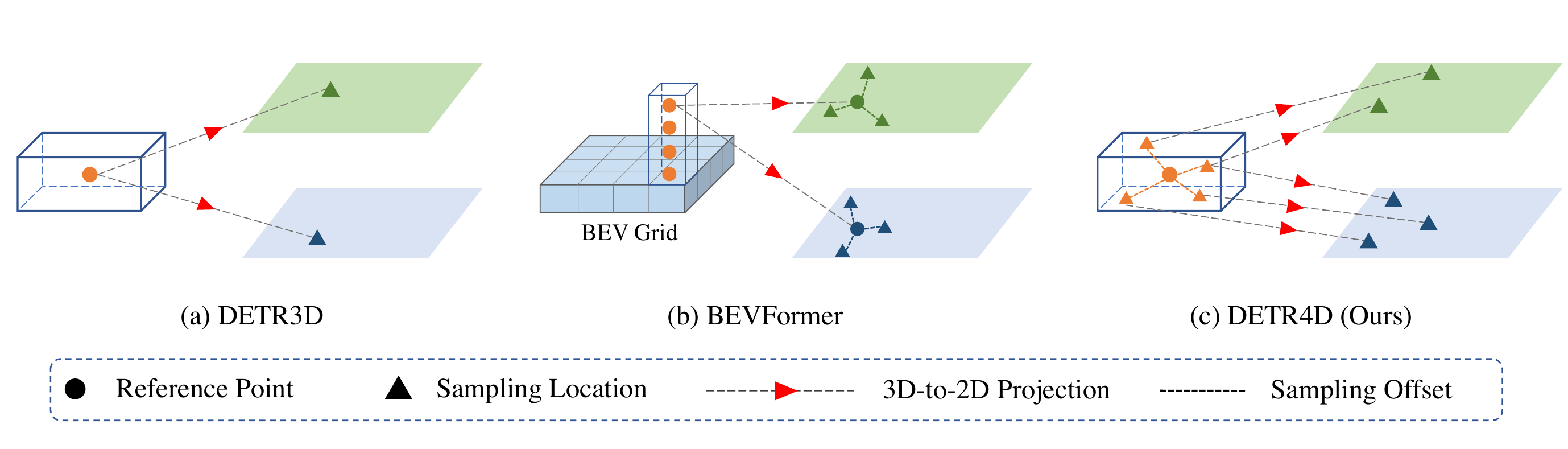}
    \vspace{-5mm}
    \caption{Comparison of different 3D-to-2D query methods. (a) DETR3D \cite{wang2022detr3d} projects object centers to images for feature sampling. (b) BEVFormer \cite{li2022bevformer} densely samples reference points in the 3D space and projects them to images. For each projected reference point, image features are aggregated using deformable attention \cite{zhu2020deformable}. (c) We propose to generate sampling locations with reference to object centers in the 3D space before projecting them to multi-view images for feature aggregation. This proposed mechanism provides enhanced geometric cues and reduces information loss for cross-view objects.}
    \label{fig:teaser}
\end{figure*}

Several studies \cite{bruls2019right, reiher2020sim2real, park2021pseudolidar,wang2022probabilistic, wang2021fcos3d} take the straightforward path of monocular detection and perform cross-camera fusion as post-processing. However, such approaches suffer from information loss for cross-view objects and result in unsatisfactory performance and efficiency \cite{wang2022detr3d}. Alternatively, a few recent works \cite{wang2022detr3d, li2022bevformer, liu2022petr} propose to treat the 3D space as a whole and construct unified representations (\textit{e.g.}, the Bird's-Eye-View (BEV) representation) from multi-view images. 
These approaches can be mainly divided into two categories, namely depth-based and query-based methods. Depth-based methods \cite{wang2019pseudo, philion2020liftlls, reading2021categorical, huang2021bevdet} back-projects 2D image features to the 3D space based on estimated pixel-wise depth distribution, which often requires extra inputs such as point clouds for depth supervision and is highly sensitive to the accuracy of depth estimation \cite{wang2022detr3d}. On the other hand, query-based approaches adopt a top-down strategy by first projecting 3D object queries (or sampling locations) onto images and then directly predicting bounding boxes from these object queries without explicitly estimating the depth.

Concretely, as the pioneering work of the query-based paradigm, DETR3D \cite{wang2022detr3d} extends the DETR \cite{carion2020enddetr} framework to 3D detection by projecting object centers to images. The projected center locations are used to sample image features for iterative query updates. Despite its simplicity and high efficiency, the performance is limited by the restricted receptive field of each object query and the error-accumulating feature sampling process due to inaccurate object locations. Follow-up works address the above issues by resorting to global attention \cite{liu2022petr} and attention-based feature construction with densely sampled points in the 3D space \cite{li2022bevformer}. While these efforts achieve remarkable performance improvements, they also bring additional computational complexity, which limits their applicability in industrial scenarios. In this work, we follow the paradigm of DETR3D to directly project 3D object queries to multi-view images for feature aggregation with sparse attention. As shown in Fig.~\ref{fig:teaser}, unlike the spatial cross-attention proposed in BEVFormer \cite{li2022bevformer} where 3D points are projected to 2D images for subsequent feature sampling, we propose a novel projective cross-attention mechanism that generates 3D sampling locations before the 3D-to-2D transformation. This proposed mechanism provides enhanced geometric cues for 3D bounding box prediction and reduces information loss for cross-view objects. Moreover, to further bridge the gap between 2D and 3D spaces, we introduce an efficient heatmap generation process through volumetric grid sampling. The heatmap prediction is used to guide the initialization of object queries by providing objectiveness priors and contextual information.

Temporal modeling has recently demonstrated great benefits in boosting the performance of camera-based 3D detection. Existing methods \cite{li2022bevformer, liu2022petrv2, huang2022bevdet4d} usually adopt the strategy of transforming intermediate spatial features with ego-motion and aggregating features from adjacent frames. However, such a strategy does not apply to our framework as direct feature query is performed on images. To this end, we propose a new hybrid temporal modeling method that temporal information is aggregated from both past object queries and image features. The proposed method introduces minimal computation overhead to the pipeline while achieving notable improvements in detection results. 
This additional temporal dimension provides rich information for 3D object detection, leading to significant performance gain. We thus name our proposed method DETR4D.

We summarize the contributions of this work as follows: 1) We propose a multi-view 3D object detection framework that follows the simple design of direct query from image features. We introduce a novel projective cross-attention mechanism to better exploit geometric cues and account for cross-view objects. Moreover, an efficient heatmap generation process is incorporated for context-aware query initialization. 2) We introduce a novel hybrid temporal modeling approach that effectively aggregates temporal information from previous object queries and image features with minimal additional computation. 3) We conduct extensive evaluations on the nuScenes dataset to validate the effectiveness of our proposed DETR4D. Experimental results demonstrate that DETR4D achieves remarkable efficiency with competitive performance.

\section{Related Works}
\label{sec:related_works}
\noindent\textbf{Transformer-based 2D Object Detection}
Transformer \cite{vaswani2017attention} is an attention-based block first proposed in natural language processing tasks, which is known for its ability for modeling long-range dependencies. Recently, a number of works \cite{liu2021swin,chu2021twins,carion2020enddetr,xie2021segformer,zhang2021meta} adapt Transformer to vision tasks and achieve remarkable progress. In particular, DETR \cite{carion2020enddetr} introduces a new end-to-end paradigm for object detection with the Transformer architecture in replacement of the heuristic designs in previous detectors \cite{ren2015faster, lin2017focal}. In DETR, object instances are represented by object queries, which interact with encoded image features with the cross-attention mechanism of Transformer for iterative updates. However, DETR suffers from slow convergence and high computation brought by the global attention operation. A number of follow-up works \cite{EfficientDETR, SparseDETR, ConditionalDETR, AnchorDETR, SMCA-DETR, DN-DETR} have been proposed to address these issues. Among them, Deformable DETR \cite{zhu2020deformable} proposes to replace the global attention in DETR with deformable attention, which is a type of sparse attention that each query of only interacts with a fixed number of sampling locations from the value. Several works \cite{li2022bevformer, zhang2021direct} extend deformable attention to 3D perception tasks by projecting 3D queries to images for feature aggregation. Our proposed projective cross-attention differs from them by predicting sampling offsets in the 3D space to exploit more instance-specific geometric cues and reduce information loss for cross-view objects.

\begin{figure*}[t]
    \centering
    \includegraphics[width=1.0\linewidth]{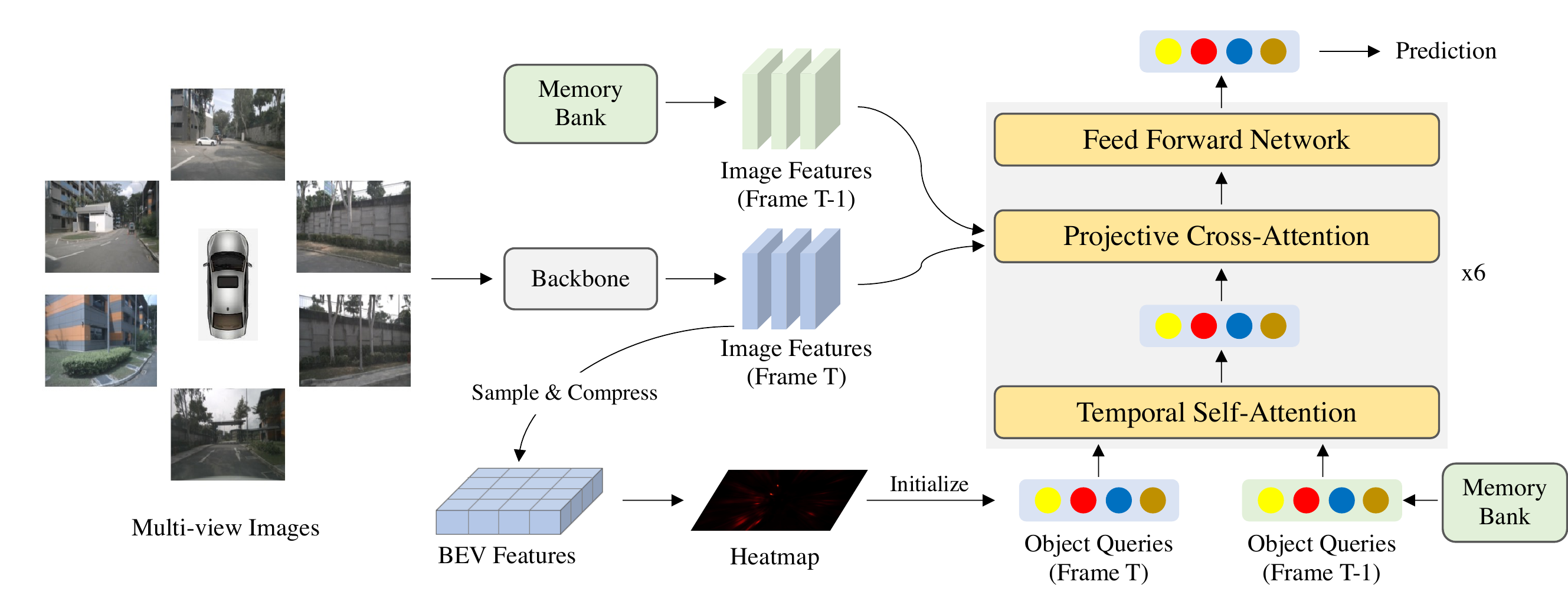}
    \vspace{-2mm}
    \caption{The architecture of DETR4D. The input multi-view images are encoded with a backbone network to produce image features. A heatmap that represents the objectiveness in the bird's-eye-view is generated to guide the initialization of object queries. The object queries directly interact with image features for efficient feature aggregation. For temporal modeling, we store past object queries and image features in a memory bank and extract useful information from them with attention mechanisms. }
    \label{fig:mdoel}
\end{figure*}

\smallskip
\noindent\textbf{Camera-based 3D Object Detection}
A number of early works \cite{chen2016monocular, brazil2019m3d, kehl2017ssd, ku2019monocular, mousavian20173d, wang2021fcos3d} perform camera-based 3D object detection in the monocular setting and largely follow 2D detection methods to generate 3D predictions in the image space. FCOS3D \cite{wang2021fcos3d} extends the 2D detector FCOS \cite{tian2019fcos} and directly predicts 3D bounding boxes from images. DD3D \cite{park2021pseudolidar} demonstrates that large-scale depth estimation pre-training can effectively improve the performance of 3D detection. Some recent works explore performing 3D detection directly in the 3D space, and they can be largely categorized into depth-based and query-based methods. Depth-based methods \cite{wang2019pseudo, philion2020liftlls, reading2021categorical, huang2021bevdet, li2022bevdepth} take the bottom-up approach to generate depth estimates for image pixels and back projects image features to the 3D space for 3D bounding box prediction. These methods often require explicit depth supervision from other input modalities such as point clouds. On the other hand, query-based methods \cite{wang2022detr3d, li2022bevformer, liu2022petr, chen2022polar, liu2022petrv2} are inspired by DETR \cite{carion2020enddetr} and object queries are used to represent objects and interact with multi-view image features with attention mechanisms. DETR3D \cite{wang2022detr3d} projects object centers to images for feature sampling. PETR \cite{liu2022petr} encodes 3D coordinates into positional embeddings and fuses them with image features. BEVFormer \cite{li2022bevformer} densely samples points over the 3D space and employs deformable attention \cite{zhu2020deformable} to generate BEV features, which the object queries interact with to generate detection predictions. Our proposed method follows the efficient design of DETR3D to directly query image features with sparse attention.  Recent studies \cite{li2022bevformer, liu2022petrv2, huang2022bevdet4d} have demonstrated the importance of temporal modeling and most existing methods perform temporal fusion with BEV features. Our proposed DETR4D provides another perspective for efficient temporal modeling by aggregating information from past queries and image features.

\section{Method}
\subsection{Overview}
Fig.~\ref{fig:mdoel} illustrates the overall architecture of the proposed DETR4D, which largely follows the structure of DETR3D \cite{wang2022detr3d}: objects are represented by queries, which are directly used to extract features from multi-view image features for iterative updates. Our proposed method differs from DETR3D with the following designs: heatmap-based query initialization (Section~\ref{sec_heatmap}), projective cross-attention (Section~\ref{sec_attention}), and temporal modeling (Section~\ref{sec_temporal}). For heatmap-based query initialization, we generate a heatmap that represents the objectiveness in the Bird's-Eye-View (BEV) through grid-based feature sampling. The predicted heatmap and the BEV features are used to guide the initialization of object queries to facilitate the subsequent interaction between queries and image features. Projective cross-attention extends from deformable attention \cite{zhu2020deformable} that it generates sampling locations based on object centers in the 3D space and projects them to images for attention-based feature aggregation. For temporal modeling, we store past object queries and image features in a memory bank during inference and extract useful information from them with attention mechanisms. After each Transformer layer, the output object queries are passed to a detection head to generate detection predictions following the pipeline of \cite{zhu2020deformable}.

\subsection{Heatmap-based Query Initialization} \label{sec_heatmap}

Existing query-based methods \cite{wang2022detr3d, liu2022petr, li2022bevformer} perform object query initialization in a random manner. However, a typical autonomous driving scene often comes with a large scale while objects only occupy a small portion of the space. A large number of object queries are usually required to ensure coverage of the entire scene (\textit{e.g.}, ~1000 queries for an input range of 100m$\times$100m), leading to redundant computations. Therefore, we are motivated to guide the initialization of queries to emphases on regions with a higher probability of object occurrence. Moreover, DETR3D \cite{wang2022detr3d} only projects object centers to images for local feature sampling, leading to a restricted receptive range and limited access to contextual information, which hinders the effectiveness of subsequent query updates. To this end, we propose a heatmap-based query initialization process to address the above issues. 

Concretely, given a set of input images, we first use an image backbone (\textit{e.g.}, ResNet \cite{he2016deepresidual}) to extract image features $F=\{F_n\}_{n=1}^{N_{view}}$, where $N_{view}$ is the number of image views. We then form a 3D sampling grid $G\in \mathbb{R}^{D\times H\times W}$ corresponding to the range of the scene and project all the sampling locations within the grid to images with the known camera transformation. We denote the projected sampling grid as $G'$. Then, we obtain the volumetric feature $F_V \in \mathbb{R}^{C \times D\times H\times W}$ with bilinear sampling and summation across views:
\begin{equation}
    F_V = \frac{1}{|\mathcal{V}_{valid}|} \sum_{n \in \mathcal{V}_{valid}} f^{bilinear}(F_n, G')
\end{equation}
where $\mathcal{V}_{valid}$ represents image views with valid projection and $C$ is the number of feature channels. We highlight that our volumetric sampling process differs from existing depth estimation-based approaches \cite{philion2020liftlls, huang2021bevdet, li2022bevdepth} by being non-parametric and more efficient. We then compress $F_V$ by concatenating along the channel dimension and encode the features with a convolution layer to obtain a flattened BEV representation $F_{BEV} \in \mathbb{R}^{C\times H\times W}$. A heatmap $M$ is generated from $F_{BEV}$ with a lightweight convolutional neural network (\textit{e.g.,} ResNet-18), which represents the objectiveness in the BEV space. To provide supervision to the predicted heatmap, we generate the ground truth heatmap $M_{gt}$ by drawing Gaussian distributions for annotated bounding boxes with a fixed radius $r$. Finally, we compute the heatmap loss with Gaussian focal loss \cite{law2018cornernetgaussianfocalloss} $L_{GF}$:
\begin{equation}
    \mathcal{L}_{heatmap} = L_{GF}(M, M_{gt}) 
\end{equation}

Subsequently, locations with high responses in the heatmap are selected as the initial locations of object queries. 
Practically, we perform non-maximum suppression with a fixed size $l$ on the heatmap before selecting top $N_{query}$ values to prevent overly dense query initialization, where $N_{query}$ is the number of object queries. The coordinates of the selected locations are encoded with a linear projection layer to form query positional encoding, and we name this process \textit{position initialization}.
To facilitate the following feature aggregation with contextual information, we sample features from $F_{BEV}$ at selected locations as the initial object queries, which we call \textit{feature initialization}.

\begin{figure}[t]
    \centering
    \includegraphics[width=0.95\linewidth]{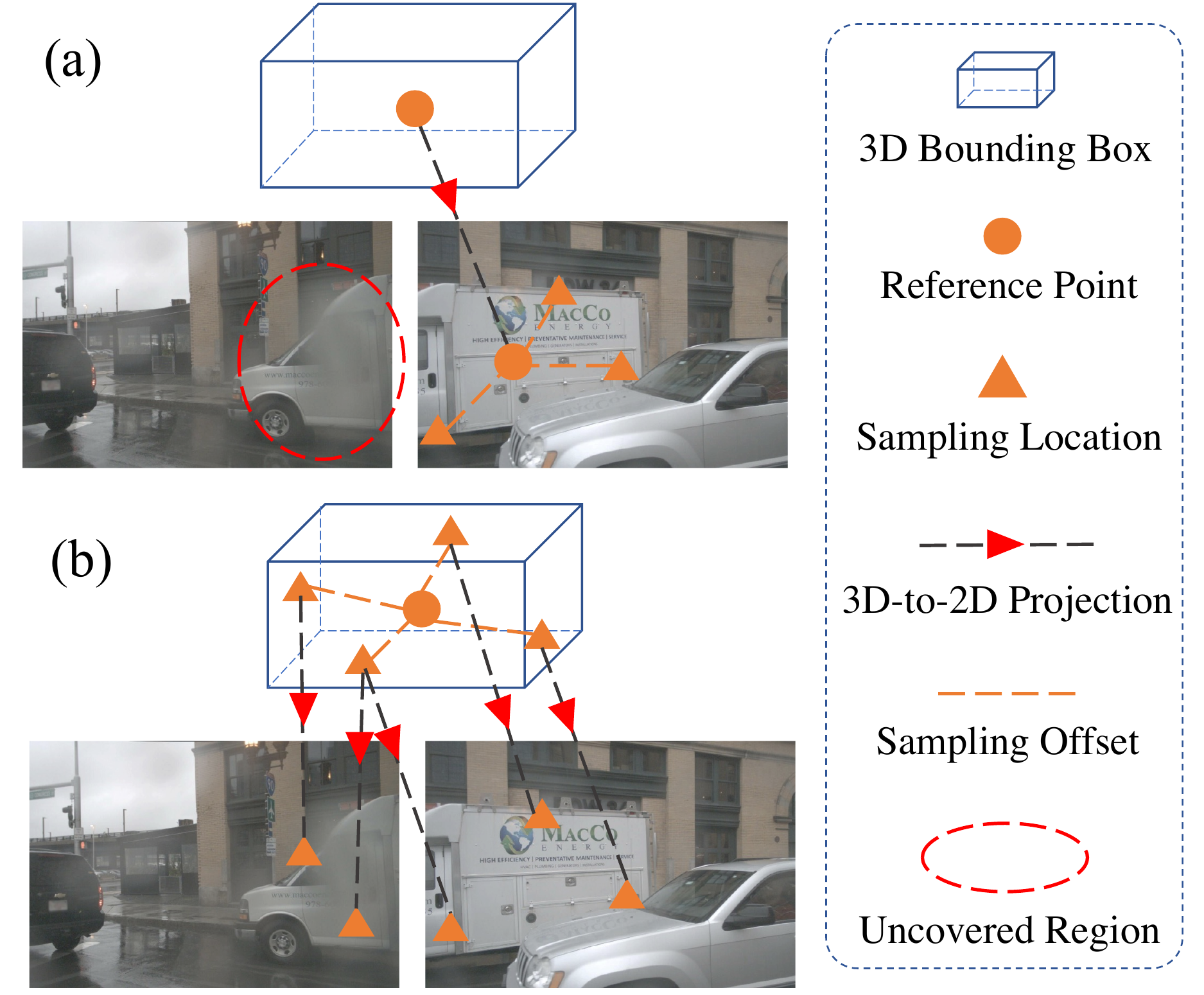}
    \caption{Comparison of the query-image interaction manner of different attention mechanisms. (a) Existing methods \cite{li2022bevformer, chen2022polar} project reference points to multi-view images and generate sampling locations in each image plane, leading to potential information loss (red circled region) for cross-view objects. (b) Our proposed projective cross-attention instead predicts sampling locations in the 3D space to provide enhanced geometric cues.}
    \label{fig:attention}
\end{figure}

\subsection{Projective Cross-Attention} \label{sec_attention}
As illustrated in Fig.~\ref{fig:teaser}(a), DETR3D \cite{wang2022detr3d} only samples image features at projected object centers, leading to a limited receptive field and poor tolerance to inaccurate object locations. Some follow-up methods \cite{li2022bevformer, chen2022polar} extend deformable attention \cite{zhu2020deformable} to 3D-to-2D query by predicting sampling offsets in each image view to adaptively aggregate features with the attention mechanism. However, such an approach still has certain restrictions when it comes to direct interaction between object queries and image features. Firstly, the projection of only object centers offers limited geometric cues that facilitate the 3D localization of objects. Secondly, as illustrated in Fig.~\ref{fig:attention}(a), it leads to potential information loss for cross-view objects when object centers only fall on individual views. To address the above issues, we propose a new projective cross-attention mechanism for the query-image interaction by extending deformable attention \cite{zhu2020deformable}.

Specifically, as shown in Fig.~\ref{fig:attention}(b), for each object query ${q}$, we first predict 3D sampling offsets $\Delta {c}_{hs}$ with respect to the object center ${c}$ by linearly projecting ${q}$, where h and s index the attention heads and sampling locations, respectively. The 3D sampling locations are then projected to image views with camera transformation to obtain corresponding image features:
\begin{equation}
    {f}_{hs} = \frac{1}{|\mathcal{V}_{valid}|} \sum_{n \in \mathcal{V}_{valid}} {W}_h f^{bilinear}(F_n, \mathcal{P}({c} + \Delta {c}_{hs}))
\end{equation}
where ${W}_h \in \mathbb{R}^{(C/N_{h})\times C}$ is the learnable weights of the value projection, $N_{h}$ is the number of attention heads, and $\mathcal{P}$ stands for the 3D-to-2D projection. The output of the projective cross-attention (PCA) is then calculated by:
\begin{equation}
    \text{PCA}(q, c, F) = \sum_h^{N_h} {W_h'} \sum_s^{N_s} A_{hs} f_{hs}
\end{equation}
where $A_{hs}$ is the attention weights generated via a linear projection based on ${q}$, and $A_{hs}$ is normalized with the Softmax operation so that $\sum_s A_{hs}=1$. $N_s$ is the number of sampling locations for each object query, and ${W}_h' \in \mathbb{R}^{C \times (C/N_h)}$ is the weights of the output projection. With the proposed attention mechanism, we incorporate instance-related geometric information into the query process to facilitate object localization. Moreover, our projective cross-attention can be easily extended to leverage multi-scale image features following the practice of deformable attention \cite{zhu2020deformable}.

\subsection{Temporal Modeling} \label{sec_temporal}
Recent studies \cite{li2022bevformer, liu2022petrv2, huang2022bevdet4d} have demonstrated the effectiveness of temporal modeling in boosting the performance in camera-based 3D detection. However, existing methods typically perform temporal fusion on intermediate spatial features (\textit{e.g.}, BEV features), which is not applicable to our proposed method where direct query-image interaction is performed. To this end, we propose a novel hybrid approach, which consists of \textit{query aggregation} and \textit{feature aggregation}, to model the temporal relationship. As illustrated in Fig.~\ref{fig:mdoel}, a memory bank is created to cache past object queries and image features. The object queries output by the last Transformer layer for the previous frame are stored. During inference, past information can be directly extracted from the memory bank to avoid repeated computation. For query aggregation, past object queries contain categorical and positional information of instances at the previous timestamp, which naturally provides useful prior knowledge to the detection task of the current frame. To pass the information to the current frame, we simply fuse object queries from both frames with the standard multi-head attention (MHA) module \cite{vaswani2017attention}. Specifically, we denote current and past queries as $Q^{(t)}$ and $Q^{(t-1)}$, respectively. We set $Q^{(t)}$ as the query of MHA and concatenate $Q^{(t)}$ and $Q^{(t-1)}$ to form key and value. Formally, our temporal self-attention (TSA) operation is defined as:
\begin{equation}
    \text{TSA} (Q^{(t)}, Q^{(t-1)}) = \text{MHA}(Q^{(t)}, [Q^{(t)}, Q^{(t-1)}])
\end{equation}
where $[\cdot]$ denotes concatenation. Note that the center location of each object is encoded as a positional encoding to the object query (see Section~\ref{sec_heatmap}), and we rectify the location of previous object queries based on ego-motion to remove the impact of self-movement before translating them into positional encodings. While object queries contain high-level instance information, we also incorporate fine-grained past image features to further improve the detection results. Our proposed projective cross-attention offers the flexibility to be easily extended to cross-frame feature aggregation. For an object query $q$ and its corresponding object center $c^{(t)}$ in the current frame, we first perform ego-motion alignment to transform the center to the previous timestamp. The transformed center location is denoted as $c^{(t-1)}$. We then perform feature aggregation from the current and previous image features in parallel and average the queried features to generate the final output. This process can be represented by:
\begin{align}
    q^{(t)} &= \text{PCA}(q, c^{(t)}, F^{(t)}) \\
    q^{(t-1)} &= \text{PCA}(q, c^{(t-1)}, F^{(t-1)}) \\
    \text{PCA}(q, c^{(t)}, F^{(t)}, &c^{(t-1)}, F^{(t-1)}) = \frac{1}{2}(q^{(t)}+q^{(t-1)})
\end{align}
Here, $F^{(t)}$ and $F^{(t-1)}$ denote image features of the current and previous frame, respectively. Thanks to the sparsity of our projective cross-attention, the proposed cross-frame feature aggregation only introduces small computation overhead to our framework. As explained, query aggregation and feature aggregation focus on different perspectives of temporal modeling and work in a complementary manner.

To train our model with temporal modeling, for each training sample at timestamp $t$, we randomly sample another frame from the past 2 seconds as the previous frame. We first run the forward pass for the previous frame to generate past object queries and image features, and gradient is not required in this process. The current frame is then input together with the past queries and features for query and feature aggregation as discussed above. During inference, a memory bank is used to cache past queries and image features to avoid repeated computation. We use a time interval of 1.5s between the previous frame and the current frame by default.

\section{Experiments}

\begin{table*}[t]
\setlength{\tabcolsep}{30pt}
\small
\caption{Performance comparison on nuScenes validation set. $\ast$ denotes initialization from a FCOS3D \cite{wang2021fcos3d} backbone. $\dagger$ denotes trained with CBGS \cite{zhu2019classcbgs}.} \label{tab:result_val} 
\vspace{-2mm}
\begin{center}\setlength{\tabcolsep}{5pt}{
\scalebox{1.0}{
\begin{tabular}{c|c|cc|ccccc}
\toprule[1.2pt]
Methods & Multi-frame & NDS$\uparrow$ & mAP$\uparrow$ & mATE$\downarrow$ & mASE$\downarrow$ & mAOE$\downarrow$ & mAVE$\downarrow$ & mAAE$\downarrow$ \\
\midrule
FCOS3D \cite{wang2021fcos3d} & & 0.295 & 0.372 & 0.806 & 0.268 & 0.511 & 1.315 & \textbf{0.170} \\
PGD \cite{wang2022probabilistic} &  & 0.409 & 0.335 & 0.732 & \textbf{0.263} & 0.423 & 1.285 & 0.172 \\
DETR3D$^{\ast}$ \cite{wang2022detr3d} &  & 0.425 & 0.346 & 0.773 & 0.268 & 0.383 & 0.842 & 0.216 \\
PolarDETR$^{\ast}$ \cite{chen2022polar} &  &  0.444 & 0.365 & 0.742 & 0.269 & \textbf{0.350} & 0.829 & 0.197 \\
PETR$\dagger$ \cite{liu2022petr}  &  & 0.442 & 0.370 & \textbf{0.711} & 0.267 & 0.383 & 0.865 & 0.201 \\
BEVFormer-S$^{\ast}$ \cite{li2022bevformer}  &  & \textbf{0.448} & 0.375 & 0.725 & 0.272 & 0.391 & \textbf{0.802} & 0.200 \\
DETR4D-S$^{\ast}$ (Ours) & & 0.444 & \textbf{0.383} & 0.719 & 0.268 & 0.415 & 0.857 & 0.213 \\
\midrule
PolarDETR-T$^{\ast}$ \cite{chen2022polar}  &  $\checkmark$ & 0.488 & 0.383 & 0.707 & \textbf{0.269} & \textbf{0.344} & 0.518 & 0.196 \\
BEVFormer$^{\ast}$ \cite{li2022bevformer} &  $\checkmark$ & \textbf{0.517} & 0.416 & \textbf{0.673} & 0.274 & 0.372 & \textbf{0.394} & 0.198 \\
DETR4D$^{\ast}$ (Ours)  & $\checkmark$ & 0.509 & \textbf{0.422} & 0.688 & \textbf{0.269} & 0.388 & 0.496 & \textbf{0.184} \\
\bottomrule[1.2pt]
\end {tabular}}}
\end{center}
\end {table*}

\begin{table*}[t]
\setlength{\tabcolsep}{30pt}
\small
\caption{Performance comparison on nuScenes test set. $\ast$ denotes initialization from a FCOS3D \cite{wang2021fcos3d} backbone. $\dagger$ denotes trained with CBGS \cite{zhu2019classcbgs}. } 
\label{tab:result_test} 
\vspace{-2mm}
\begin{center}\setlength{\tabcolsep}{5pt}{
\scalebox{1.0}{
\begin{tabular}{c|c|cc|ccccc}
\toprule[1.2pt]
Methods & Multi-Frame & NDS$\uparrow$ & mAP$\uparrow$ & mATE$\downarrow$ & mASE$\downarrow$ & mAOE$\downarrow$ & mAVE$\downarrow$ & mAAE$\downarrow$ \\
\midrule
FCOS3D \cite{wang2021fcos3d} & & 0.428 & 0.358 & 0.690 & 0.249 & 0.452 & 1.434 & \textbf{0.124} \\
PGD \cite{wang2022probabilistic} &  & 0.448 & 0.386 & \textbf{0.626} & \textbf{0.245} & 0.451 & 1.509 & 0.127 \\
PETR$\dagger$ \cite{liu2022petr}  & & 0.455 & 0.391 & 0.647 & 0.251 & \textbf{0.433} & 0.933 & 0.143 \\
BEVFormer-S$^{\ast}$ \cite{li2022bevformer} & & 0.462 & \textbf{0.409} & 0.650 & 0.261 & 0.439 & 0.925 & 0.147  \\
DETR4D-S$^{\ast}$ (Ours) & & \textbf{0.463} & 0.406 & 0.684 & 0.256 & 0.435 & \textbf{0.873} & 0.152 \\
\midrule
BEVFormer$^{\ast}$ \cite{li2022bevformer}  & $\checkmark$  & \textbf{0.535} & 0.445 & \textbf{0.631} & 0.257 & \textbf{0.405} & \textbf{0.435} & 0.143 \\
DETR4D$^{\ast}$ (Ours) & $\checkmark$  & 0.530 & \textbf{0.452} & 0.645 & \textbf{0.251} & 0.419 & 0.507 & \textbf{0.136} \\
\bottomrule[1.2pt]
\end {tabular}}}
\end{center}
\end {table*}

\subsection{Datasets}
We conduct experiments on nuScenes \cite{caesar2020nuscenes} for evaluation. The nuScenes dataset consists of 1000 sequences, and each sequence has a duration of around 20 seconds. Each sample in a sequence consists of 6 images collected by cameras facing different directions, and the combination of all images covers the $360\degree$ FOV. The dataset is officially divided into training, validation, and testing datasets with 700, 150, and 150 sequences, respectively. Annotations are provided every 0.5 seconds and annotated samples are referred to as keyframes. We only use keyframes in our experiments. We use the official evaluation metrics, including mean Average Precision (mAP), mean Average Translation Error (mATE), mean Average Scale Error (mASE), mean Average Orientation Error(mAOE), mean Average Velocity Error(mAVE), mean Average Attribute Error(mAAE), as well as nuScenes Detection Score (NDS).

\subsection{Implementation Details}
We follow previous methods \cite{wang2022detr3d, li2022bevformer} and use ResNet-101 \cite{he2016deepresidual} initialized from FCOS3D \cite{wang2021fcos3d} as our backbone. A FPN \cite{lin2017featurefpn} takes as input the output of the backbone model and generates multi-scale features with sizes 1/8, 1/16, 1/32, and 1/64 of the input size. We use a default image input size of 1600 $\times$ 640 and a perception range of [-51,2m, 51.2m] for the X-Y plane and [-5m, 3m] for the Z axis. For heatmap generation, a grid shape of (144, 144, 8) is used for the X, Y, and Z axes. We set both $r$ and $l$ equivalent to 3 pixels on the heatmap. We use 8 attention heads for all Transformer attention modules and 8 sampling points for projective cross-attention. We follow existing methods \cite{li2022bevformer, liu2022petr} and use 900 object queries by default. The model is trained for 24 epochs with AdamW \cite{loshchilov2017decoupledadamw} optimizer and a learning rate of $2 \times 10^{-4}$. We train the model with a batch size of 8 on 8 NVIDIA V100 GPUs.

\subsection{Benchmarking Results}
We compare our model with existing camera-only detection methods on the nuScenes dataset. For a fair comparison, we do not include methods that use additional point cloud data for explicit depth supervision. To study the effectiveness of temporal modeling, we also train our model with single-frame input and denote it as DETR4D-S. Tab.~\ref{tab:result_val} and Tab.~\ref{tab:result_test} report results on the validation and test sets, respectively. DETR4D obtains comparable performance to the state-of-the-art method BEVFormer \cite{li2022bevformer} on both sets despite adopting the simple design of direct query-image interaction. As compared with the base model DETR3D \cite{wang2022detr3d}, our single-frame version DETR4D-S achieves significantly improved detection results including an increase of 3.7 \% in mAP on the validation set. With the proposed temporal modeling based on multi-frame input, DETR4D further brings a substantial performance boost on top of DETR4D-S consistently on both sets.

\begin{table}[t]
\setlength{\tabcolsep}{30pt}
\small
\caption{Effect of heatmap-based query initialization. Position Init.: initialize query positions with heatmap. Feature Init.: initialize query features from $F_{BEV}$.} \label{tab:ablation_heatmap} 
\vspace{-2mm}
\begin{center}\setlength{\tabcolsep}{5pt}{
\scalebox{1.0}{
\begin{tabular}{cc|cc}
\toprule[1.2pt]
Position Init. & Feature Init. & NDS$\uparrow$ & mAP$\uparrow$ \\
\midrule
& & 0.417 & 0.348 \\
$\checkmark$ & &  0.429 & 0.362 \\
$\checkmark$ & $\checkmark$ & \textbf{0.439} & \textbf{0.367} \\
\bottomrule[1.2pt]
\end {tabular}}}
\vspace{-2mm}
\end{center}
\end {table}

\begin{figure}[t]
    \centering
    \includegraphics[width=1.0\linewidth]{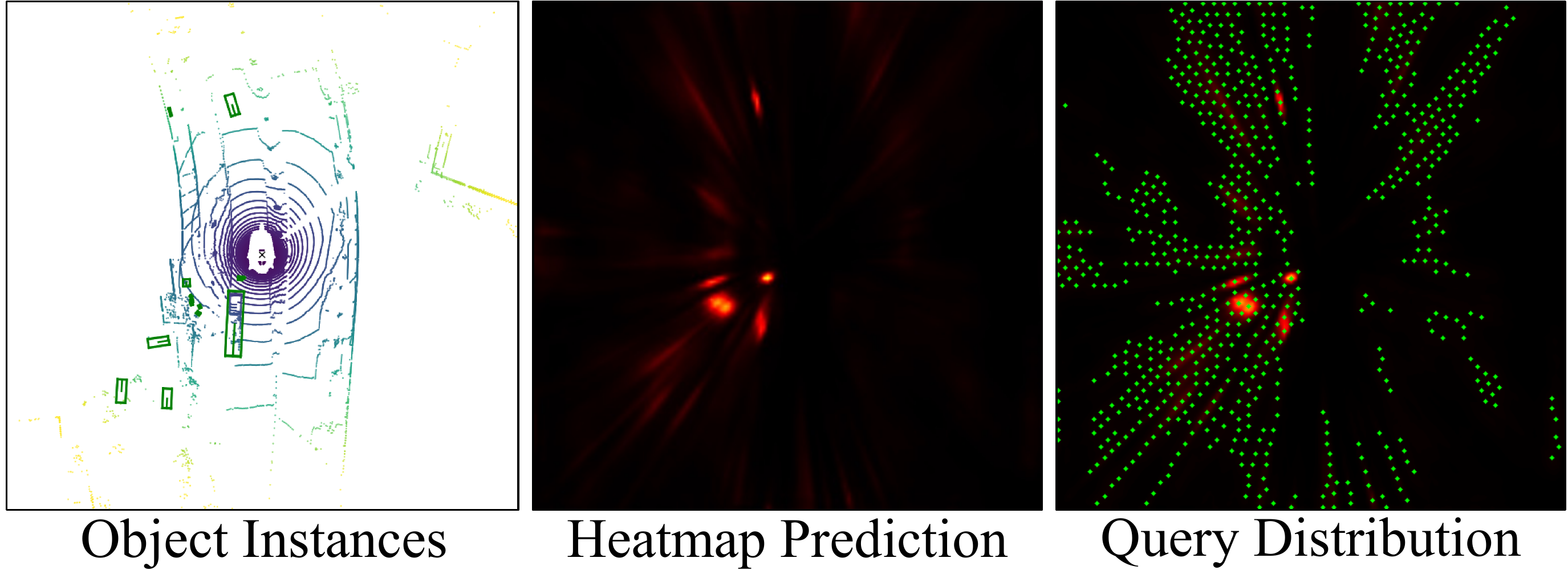}
    \caption{Visualization of heatmap-based query initialization. Object queries are initialized at regions with a high likelihood of being the foreground. Green points indicate initial query positions.}
    \label{fig:heatmap}
\end{figure}

\subsection{Ablation Study}
We conduct experiments to study the effect of the model components. Results are reported on the validation set of the nuScenes dataset. To reduce training time, we conduct all the experiments with a reduced input size of $1280 \times 512$ unless otherwise stated. 

\begin{table}[t]
\setlength{\tabcolsep}{30pt}
\small
\caption{Effect of projective cross-attention.} \label{tab:ablation_pca} 
\vspace{-2mm}
\begin{center}\setlength{\tabcolsep}{5pt}{
\scalebox{1.0}{
\begin{tabular}{c|cc}
\toprule[1.2pt]
Methods & NDS$\uparrow$ & mAP$\uparrow$ \\
\midrule
Spatial Cross-Attention \cite{li2022bevformer} & 0.425 & 0.358 \\
Projective Cross-Attenion (Ours) & \textbf{0.439} & \textbf{0.367} \\
\bottomrule[1.2pt]
\end {tabular}}}
\end{center}
\end {table}

\smallskip
\noindent\textbf{Effect of heatmap-based query initialization.}
Our proposed heatmap-based query initialization consists of position initialization and feature initialization. The former provides prior knowledge of objectiveness, while the latter equips object queries with contextual information. We conduct experiments with our single-frame model to study their effects and report the results in Tab.~\ref{tab:ablation_heatmap}. We can observe that each of them plays a positive role in improving the detection performance, which validates the effectiveness of our proposed query initialization strategy. We also visualize the initialization process in Fig.~\ref{fig:heatmap}. It can be seen that queries are initialized at regions with higher probability of being the foreground. 

\smallskip
\noindent\textbf{Effect of projective cross-attention (PCA).} 
To study the effect of the proposed PCA, we replace it with the spatial cross-attention mechanism proposed in \cite{li2022bevformer}, which is also an extension from deformable attention \cite{zhu2020deformable} but generates sampling locations in 2D images. As shown in Tab.~\ref{tab:ablation_pca}, PCA brings a notable margin in performance over the comparing method as it provides enhanced geometric cues and reduces information loss for cross-view objects.

\begin{table}[t]
\setlength{\tabcolsep}{30pt}
\small
\caption{Effect of temporal modeling. Query Aggr.: temporal fusion with past object queries. Feature Aggr.: temporal fusion with past image features.} \label{tab:ablation_temporal}
\vspace{-2mm}
\begin{center}\setlength{\tabcolsep}{5pt}{
\scalebox{1.0}{
\begin{tabular}{cc|cc}
\toprule[1.2pt]
Query Aggr. & Feature Aggr. & NDS$\uparrow$ & mAP$\uparrow$ \\
\midrule
 & & 0.439 & 0.367 \\
 $\checkmark$ & &  0.480 & 0.399 \\
 &  $\checkmark$ & 0.497 & 0.407 \\
 $\checkmark$ &  $\checkmark$ & \textbf{0.505} & \textbf{0.420} \\
\bottomrule[1.2pt]
\end {tabular}}}
\end{center}
\end {table}

\begin{table}[t]
\setlength{\tabcolsep}{30pt}
\small
\caption{Effect of ego-motion alignment in temporal modeling.} \label{tab:ablation_ego} 
\vspace{-2mm}
\begin{center}\setlength{\tabcolsep}{5pt}{
\scalebox{1.0}{
\begin{tabular}{c|cc}
\toprule[1.2pt]
Ego-motion Alignment & NDS$\uparrow$ & mAP$\uparrow$ \\
\midrule
& 0.466 & 0.388 \\
$\checkmark$ & \textbf{0.505} & \textbf{0.420} \\
\bottomrule[1.2pt]
\end {tabular}}}
\end{center}
\end {table}

\smallskip
\noindent\textbf{Effect of temporal modeling.}
Our proposed temporal modeling adopts a hybrid approach that combines query aggregation based on past object queries and feature aggregation based on past image features. We study the effect of both components with experiments. As reported in Tab.~\ref{tab:ablation_temporal}, each aggregation method alone introduces a substantial performance improvement over the single-frame baseline, while they are complementary to each other that the best performance is obtained when both are applied. The results validate our analysis in Section~\ref{sec_temporal} that query and feature aggregations address different aspects of temporal modeling.

\begin{figure*}[t]
    \centering
    \includegraphics[width=1.0\linewidth]{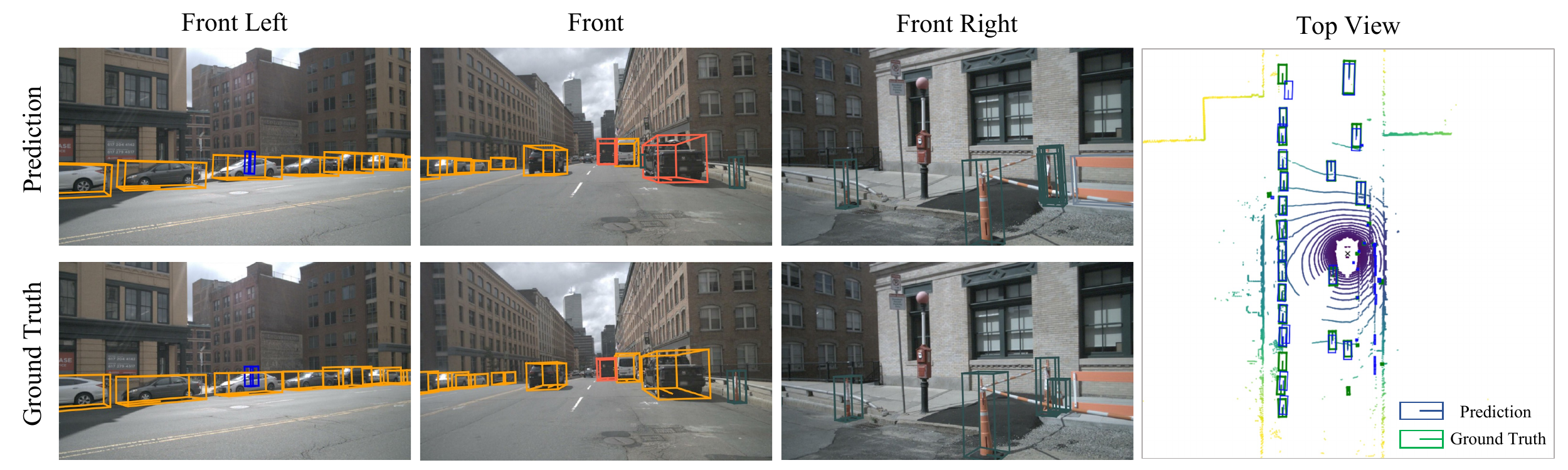}
    \vspace{-2mm}
    \caption{Qualitative results of DETR4D. We show three front camera views and the top view for illustration.}
    \label{fig:visual}
\end{figure*}

\smallskip
\noindent\textbf{Effect of ego-motion alignment.}
In temporal modeling, we perform ego-motion alignment to object locations to rule out the impact of self-movement. We can see from Tab.~\ref{tab:ablation_ego} that the absence of ego-motion alignment causes a steep degeneration in model performance, which demonstrates its importance and necessity.

\smallskip
\noindent\textbf{Time interval between frames.}
We study the impact of the time interval between previous and current frames. As shown in Tab.~\ref{tab:ablation_interval}, the model performance improves as we increase the interval from 0.5s to 1.5s, but decreases as we further enlarge the interval. This behavior is expected as a larger interval leads to more distinct features from adjacent frames and thus potentially provides more meaningful information. Moreover, adjacent frames can also be viewed as a binocular system that forms a stereo, which leads to more accurate depth estimation. However, when the interval becomes too large, it makes the cross-frame association difficult and causes a performance drop.

\begin{table}[t]
\setlength{\tabcolsep}{30pt}
\small
\caption{Time interval between the previous and current frames for temporal modeling.} \label{tab:ablation_interval} 
\vspace{-2mm}
\begin{center}\setlength{\tabcolsep}{5pt}{
\scalebox{1.0}{
\begin{tabular}{c|cc}
\toprule[1.2pt]
Interval & NDS$\uparrow$ & mAP$\uparrow$ \\
\midrule
0.5s & 0.486 & 0.403\\ 
1.0s & 0.503 & 0.416 \\
1.5s & \textbf{0.505} & \textbf{0.420} \\ 
2.0s & 0.502 & 0.418 \\
\bottomrule[1.2pt]
\end {tabular}}}
\end{center}
\end {table}

\begin{figure}[t]
    \centering
    \includegraphics[width=1.0\linewidth]{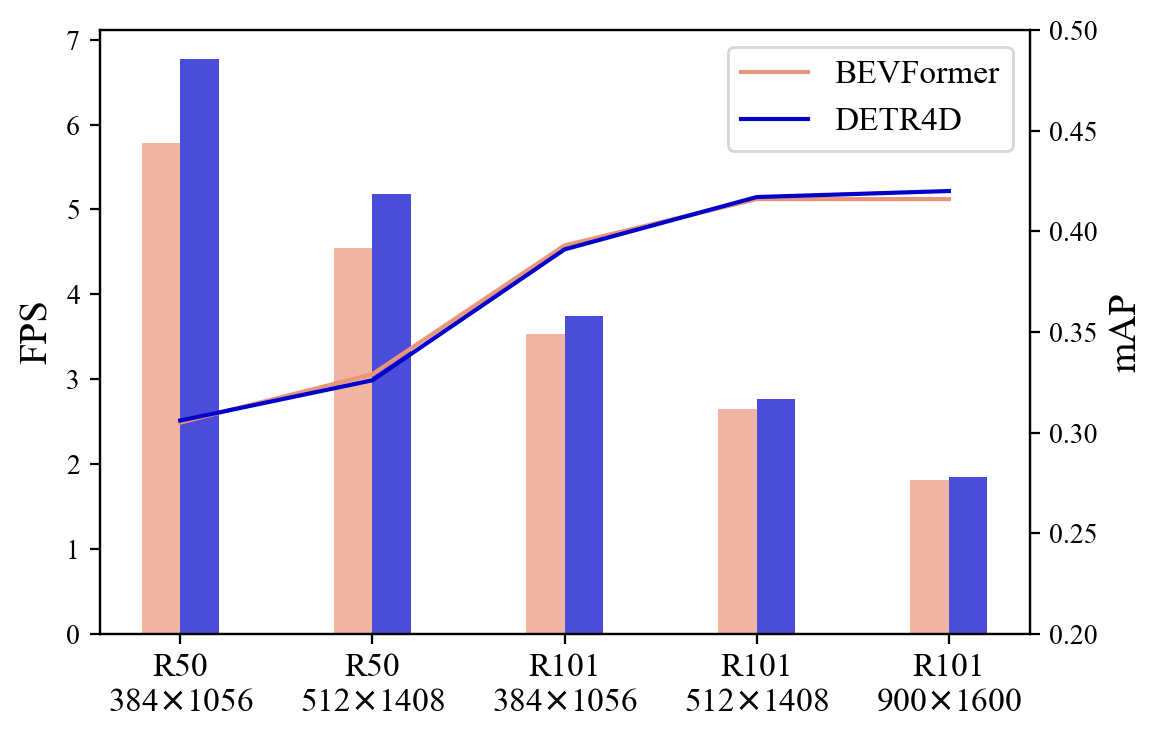}
    \vspace{-2mm}
    \caption{Inference speed comparison. Bars indicate Frame Per Second (FPS) and lines indicate mean Average Precision (mAP). ResNet-50 (R50) and ResNet-101 (R101) backbones are applied with different input image sizes.}
    \label{fig:speed}
\end{figure}

\smallskip
\noindent\textbf{Inference speed.}
We evaluate the inference speed with different backbones and compare our method with the state-of-the-art query-based method BEVFormer \cite{li2022bevformer}. As shown in Fig.\ref{fig:speed}, DETR4D achieves higher efficiency while maintaining comparable performance. The difference in inference speed is larger with a smaller backbone model or input size because the computation is dominated by the image backbone when a large backbone is applied.

\subsection{Qualitative Results}
We visualize the detection results of a sample frame in Fig.~\ref{fig:visual}. DETR4D achieves satisfactory results for most objects while making a few mistakes for small or distant instances.

\section{Conclusion}
In this work, we propose DETR4D, a simple and efficient query-based method for multi-view 3D detection. Our model skips the generation of intermediate features and directly predicts detection results by aggregating features from images. We introduce a novel projective cross-attention module for enhanced geometric information exploitation and an efficient heatmap generation process for guided query initialization. Moreover, we provide a new perspective for temporal modeling by introducing a hybrid approach that extracts information from both past object queries and image features. Extensive experiments show that DETR4D achieves remarkable efficiency and competitive performance. 

\smallskip
\noindent\textbf{Limitations.} 
As compared to other data modalities such as point clouds, images have the disadvantages of lacking depth information and being sensitive to environmental conditions such as weather and lighting. As a camera-based approach, our method is also affected by these aspects. To address this limitation, multi-modal methods could be further studied in future research.

{\small
\bibliographystyle{ieee_fullname}
\bibliography{egbib}
}

\end{document}